# How Case-Based Reasoning Explains Neural Networks: A Theoretical Analysis of XAI Using *Post-Hoc* Explanation-by-Example from a Survey of ANN-CBR Twin-Systems

Mark T. Keane[1,2,3] and Eoin M. Kenny[1,2]

[1] School of Computer Science, University College Dublin, Dublin, Ireland
[2] Insight Centre for Data Analytics, University College Dublin, Dublin, Ireland
[3] VistaMilk SFI Research Centre, University College Dublin, Dublin, Ireland
{mark.keane, eoin.kenny}@insight-centre.org

**Abstract.** This paper proposes a theoretical analysis of one approach to the eXplainable AI (XAI) problem, using *post-hoc* explanation-by-example, that relies on the twinning of artificial neural networks (ANNs) with case-based reasoning (CBR) systems; so-called ANN-CBR twins. It surveys these systems to advance a new theoretical interpretation of previous work and define a road map for CBR's further role in XAI. A systematic survey of 1102 papers was conducted to identify a fragmented literature on this topic and trace its influence to more recent work involving deep neural networks (DNNs). The twin-system approach is advanced as one possible coherent, generic solution to the XAI problem. The paper concludes by road-mapping future directions for this XAI solution, considering (i) further tests of feature-weighting techniques, (ii) how explanatory cases might be deployed (e.g., in counterfactuals, *a fortori* cases), and (iii) the unwelcome, much-ignored issue of user evaluation.

## 1 Introduction

As AI systems impact our everyday lives, jobs, and leisure time, the issue of explaining how these systems actually work has become more acute, the so-called eXplainable AI (XAI) problem. In the last few years, almost every major AI/ML conference has targeted this problem either as a major theme or as a focus for thematic workshops (e.g., NIPS-16, IJCAI-17, IJCAI/ECAI-18, IJCNN-17, ICCBR-18, ICCBR-19) along with the emergence of meetings dedicated solely to it (FAT-ML, FAT*19; see [1]). The urgency of this effort is not just academic, as it is played out against a backdrop of increasing regulatory interest by governments in AI (e.g., GDPR in the EU [2, 3]). This paper surveys one particular solution to the XAI problem, where an opaque, black-box AI system is explained by a more interpretable, white-box AI system; the so-called *twin-systems* approach [4]. This survey is used to advance a new theoretical interpreta-tion of previous work and define a road map for CBR's further role in XAI. Specifically, we review the pairing of artificial neural networks[1] (ANNs) with case-based reasoning (CBR) systems where the explanatory cases of the latter are used to interpret the opaque outputs of the former; so-called ANN-CBR twins. For example, imagine an ANN that accurately predicts house prices given different feature-descriptions of houses (e.g., *size*, *location*, *no-of-rooms*), but like all ANNs is opaque and, thus, cannot explain its predictions. If we twin this ANN with a CBR system, using the same dataset, extract the ANN's feature-weights and apply them in the CBR-system to retrieve neighboring cases to the query-case, then we can use the latter to explain the predictions of the former (see Fig. 1).

We have discovered a fragmented literature on this topic that deserves to be brought together, if only to avoid unnecessary re-invention. In the next sub-section, we lay out our orientation to "explanation" and the motivation for the present systematic survey.

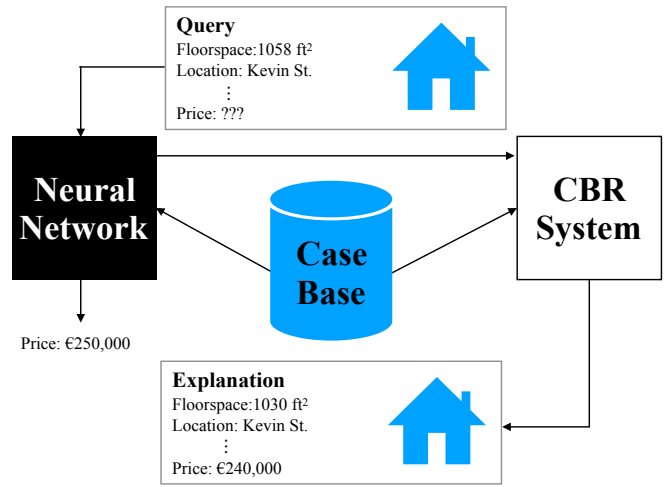

**Fig. 1**: A simple ANN-CBR twin-system (adapted from [Kenny and Keane [4]]; a query-case posed to an ANN gives an accurate, but unexplained, prediction for a house price. The ANN is twinned with the CBR system (both use the same dataset), and its feature-weights are analyzed and used in the CBR system to retrieve nearest-neighbor cases to explain the ANN's prediction.

## 1.1 "Explanation" Needs Explanation

As an area, XAI has many issues; foremost amongst these, perhaps, is some clarity on what "explanation" actually means. Several recent XAI reviews have pointedly noted the lack of clear definitions for the notions of explanation, interpretability and transparency [5-10], echoing long-standing discussions in CBR [11, 12], recommender systems [13], Philosophy [14-16] and Psychology [17]. While the exact meaning of these terms remains a matter of debate, these reviews make useful taxonomic distinctions. For example, Sørmo et al.'s [11] review reports the distinction between explaining how the system reached some answer (which they call *transparency*) and explaining why the answer is good (*justification*). More recently, this distinction is echoed by dividing *interpretability* into (i) *transparency* (or *simulatability*) which tries to reflect *how* the AI system produced its outputs, and (ii) *post-hoc interpretability* which is more about *why* the AI reached its outputs, providing some after-the-fact rationale/evidence for system outputs [7]. CBR systems are notable in this respect, through their use of examples/cases/precedents to explain system outputs [7, 10, 11, 12, 18, 19]. So, here, when a CBR system's cases are used to explain an ANN's opaque predictions, it is classed as "*post-hoc* explanation-by-example". As such, twin-systems are just one possible solution to interpretability in the XAI problem, but one, we argue, that deserves more attention.

---

[1]  Here, ANN is used to label all neural network techniques; older neural networks will be labelled as multi-layered perceptrons (MLP) and newer deep learning techniques will be called "deep neural networks" (DNNs; following [5]).

### 1.2 Motivation for a Systematic Review

There are several reasons why a systematic review of ANN-CBR twins for XAI is both timely and necessary. First, if citation patterns are any indication, there is clear evidence that the literature on twin-systems is fragmented. For instance, many recent reviews of XAI make little or no reference to key twin-systems papers in the CBR literature [1, 6, 7], while referencing papers outside CBR canon [20, 21]. Second, if we do not know the literature on this CBR-solution to XAI then, arguably, we are doomed to re-invent its findings and mistakes. Third, the XAI area requires a systematic, general framework to bring the literature together and focus future efforts. As Pedreschi et al. [22] point out "the state of the art to date still exhibits *ad-hoc*, scattered results, mostly hardwired to specific models…[and]… a widely applicable, systematic approach has not emerged yet". The twin-system idea represents one possible general solution to a broad class of systems. Fourth, a systematic survey should allow us to know where we currently stand, and then to strategically road map future directions for this XAI solution. Hence, in the remainder of this paper we review the literature on ANN-CBR twins as solutions to the XAI problem. In Section 2, twin-systems are defined more precisely. In Section 3, our systematic survey methodology is described using the twin-system definition along with descriptive statistics. In Section 4, using the literature found, the history of ANN-CBR twins is outlined. Finally, in Section 5, future directions are road mapped.

## 2. Defining ANN-CBR Twins

ANN-CBR twin-systems can be found at the intersection between research on ANNs [23, 24], CBR [25-27], and hybrid systems [28-30] when explanation is a major task-requirement of the system.

**Artificial Neural Networks (ANNs).** Biologically inspired, these AI systems typically consist of layers of nodes with non-linear activation functions and a bias term, which are connected by weights [23, 24]. Here, we distinguish between traditional neural networks of the multilayer perceptron (MLP) or backpropagation (BP) variety, and deep neural networks (DNNs) which include a wide variety of techniques; such as, recurrent neural networks (RNNs), convolutional neural networks (CNNs) and generative adversarial networks (GANs) [31-34]. MLPs typically consist of three layers, an input feature layer, a hidden layer (aka, its latent features), and an output layer. At their simplest, these ANNs learn an input-output mapping over a training set, so that when a test-case is presented, its features are used to accurately predict/classify at the MLP output layer.

Significantly, the model's learning of an input-output mapping depends on modifying the weights connecting the nodes in these layers and the bias terms within the nodes. DNNs are a menagerie of many different techniques; notably, they advance beyond MLPs by being able to learn features in unstructured data (such as images or video). However, the non-linear nature of all of these ANNs make them difficult to interpret and poor at explaining their outputs [35-37]. Attempts to make ANNs more interpretable use many different "explanation methods" that are often specific to a given architecture (see reviews [5, 9, 37]). Arguably, DNNs are even less interpretable than MLPs, because of their complexity and difficulties in surfacing their extracted features. Currently, major efforts at "explaining" DNNs hinge on visualizing what specific neurons have learned or indicating "where the DNN is looking" in an image using saliency maps [36, 38-40]. However, these methods are often quite specific to particular DNN-techniques and do not reflect the model's "reasoning process" [83]. So, a key question for the field is whether any approach can explain all ANNs – both MLPs and DNNs – in a general, unified way [5]; arguably, ANN-CBR twins are one possible solution [4].

**Case-Based Reasoning (CBR).** These systems preform a type of reasoning from examples or cases using a *retrieval*, *reuse*, *revise*, and *re-train* cycle [25-27, 41]. At its simplest, in CBR, when a query-case is presented the most similar cases to it are retrieved before being adapted (or used directly) to make a prediction/classification. Typically, the *retrieval* step finds cases by matching the features of the query-case and cases in the case base using $k$-nearest neighbor ($k$-NN). Retrieval accuracy (and, hence, the success of the system) can depend heavily on the weights given to these features, weights that reflect their importance in the domain. Notably, CBR is claimed to have a "natural" transparency as its reasoning-from-precedent or -example parallels what human experts sometimes do [18, 25]; though these claims have not always been extensively user-tested [11]. Accordingly, CBR as an area has a substantial literature on explanation [11, 12, 19], as does its sister area of recommender systems [13, 42].

## 2.1 ANN-CBR Twins

ANN-CBR twins are a special-case of a hybrid system, that combines ANN and CBR modules, when both accuracy and interpretability are primary requirements of the overall system. Though ANNs and CBR were coupled as early hybrid systems [43-45], it is not really until the late-1990s that "true" *twins* emerge [20, 46-54]. Fig. 1 shows one simple example of an ANN-CBR twinning. The task, here, is the prediction of house prices, where one has some dataset of training examples (i.e., a case base of prior cases) describing houses and their prices from previous years. The ANN accurately learns to predict the price of unseen houses (i.e., query-cases) having computed the input-output mapping from house-features to their price using the training set. To explain the ANN's prediction, its feature-weights are (in some way) extracted and used in the CBR's $k$-NN retrieval-step, to identify a nearest-neighbor case (or cases) to "explain" the ANN's prediction. In essence, the explanation step is asserting: price-$x$ is predicted, because these other houses, that have very similar features, have these prices (that are close to the predicted one). Of course, the success of this whole enterprise depends on a number of factors: (i) the ANN has to be reasonably accurate in its predictions, (ii) the feature-weights extracted from the ANN have a high fidelity to the ANN's function (iii) the nearest neighbors found do not bear an overly complex relationship to the query-case (iv) and the user has sufficient expertise to easily relate these explanatory nearest neighbors to the query case (e.g., as in Fig 1 and Fig. 2; see also [11]) and so on.

**Definition of ANN-CBR Twin.** Accordingly, we can define an ANN-CBR twin-system, precisely, as a system with:

- *Two Techniques*. A hybrid system in which (at least) two techniques[2] – ANN (MLP or DNN) and CBR techniques (notably, $k$-NN) – are combined to meet system requirements of accuracy and interpretability.

- *Separate Modules*. Where these techniques are run as separate modules, independently but, as it were, side-by-side.

- *Common Dataset*. The two techniques are run on the same dataset (i.e., they are twinned by this common usage).

- *Feature-Weight Mapping.* Some description of the ANN's functionality, typically described as its *feature-weights*, that "reflect" what the ANN has learned, is mapped to the $k$-NN retrieval step of the CBR-system.

- *Bipartite Division of Labor*. In the ANN and CBR modules, the former delivers predictions and the latter provides interpretability, explaining the ANN's outputs (for classification or regression).

As we shall see in our subsequent survey, though this is quite a simple definition, it excludes many hybrid systems that combine ANNs and CBR, as well as many CBR systems that do explanation *without* any ANN-aspect. For example, there are many systems that combine ANNs and CBR in a *pipelined* way where the ANN is used to extract features or feature weights that then improve the CBR's performance, using both MLPs [55-57] and DNNs [58]; these are *not* twin-systems because the CBR module is making the predictions (though the ANN improves these predictions), and the CBR system is not explaining the ANN's predictions. Similarly, there are some systems that use ANN and CBR modules in a single system, where *both* make predictions [59-61]. For instance, several agent-based systems for predicting oceanographic events (e.g., sea temperature, oils spills, red tides) alternate between ANN and CBR sub-systems, where the predictions from both are monitored to ensure continuing accuracy over time [59-61]; here, both systems are tasked with accuracy and the CBR sub-system is not specifically tasked with explanation (i.e., there is no mapping of feature weights). In the next section, we survey the somewhat abandoned regions of the hybrid-systems literature, relating to *true* ANN-CBR twins that specifically address explanation.

## 3    A Systematic Review: Methodology

A systematic search of the literature on ANN-CBR twins for explanation was done with a number of top-down searches using relevant keywords, supplemented by bottom-up, citation-based searches from key papers (see Table 1). In total 1,102 papers were checked (title and abstract) and filtered down to 379 papers; from this latter set a close reading of 90 papers was carried out to identify all the ANN-CBR twins in the literature.

---

2    Note, there are many other systems that combine CBR with other techniques, that are not considered here (e.g., with Genetic Algorithms, Rule-Based systems, Bayesian techniques).

### 3.1 Method: Search Procedure

Five systematic searches were carried out on https://scholar.google.com between January $6^{th}$, 2019 and March $24^{th}$, 2019: four top-down searches using keywords and one bottom-up search through papers that cited key articles. Table 1 shows the string searches used in each of the top-down searches with (i) the number of results considered for a given search (these results were checked by reading the title and abstract, and checking the google anchor-text on which the strings matched), (ii) the unique papers selected across these searches that were considered further (N = 379; these papers were generally downloaded and string-searched for the part of the paper that discussed their hybrid status and "explanation"). From the latter set, a final set of papers (N=90) were selected to be read in full to determine if the paper described a twin-system, as defined. In all searches, review papers were excluded as we wanted original system papers. We also tried not to double-count cases of twin-systems; for example, groups often produce several papers in different venues for the same system, so where the papers were essentially identical we did not double count them (though we did count cases where essentially the same system was applied to a new domain/task).

**Table 1.** Five distinct searches performed in GoogleScholar (Jan-Mar, 2019) showing the keywords used, the number of results checked and the total number of unique papers that were identified for closer reading.

| Search Terms | # | Paper-Results Relevance Checked | Unique Papers Selected For Reading |
|---|---|---|---|
| "hybrid systems for explanation" "survey" "review" | 1 | 200 | 12 |
| "hybrid" "CBR" "explain" "explanation" | 2 | 250 | 211 |
| "ANN" "CBR" "explanation" | 3 | 100 | 79 |
| "NN" "Neural Network" "CBR" "explanation" | 4 | 200 | 57 |
| None (Manual check of citations to 8 key papers) | 5 | 352 | 20 |
| *Totals* | | 1,102 | 379 |

### 3.2 Results Summary

In total 1,102 distinct GoogleScholar results were initially checked and filtered down to 379 potentially relevant papers, that were downloaded and examined for evidence of being twin-systems. From this set, only 90 were read in full to see if they match the twin-system definition. Of these 90 papers, only 34 were identified as *true* ANN-CBR twin-systems (n.b., only 21 of these papers report unique systems). Many systems combine ANN and CBR techniques, but fail to meet some key property of the twin-system definition (e.g., they were a pipeline, they did not work over the same dataset, or explanation was not a major concern). There was some indication that the top-down searches identified a fairly complete set of relevant papers because (i) many of the same papers recurred across searches, (ii) several, apparently, plagiarized papers were found, where the same paper was published with non-overlapping author names [62, 63], and (iii) many of the papers found in top-down searches cited the key papers on the twinning topic in the bottom-up search. The character and profile of the identified papers is discussed as a history in the next section.

# 4    A History of ANN-CBR Twins

Our survey of the landscape of ANN-CBR twins reveals a fragmentary and subdued development of the twinning idea. A close reading of 90 articles on hybrid ANN-CBR systems that made some mention of explanation, found only 34 papers (21 unique systems) that were *true* twin-systems. The remaining papers tend to be ANN-CBR pipelines where the ANN is used to compute features and/or feature-weights that are then used in the CBR's $k$-NN for predictive purposes. Even though the idea of combining ANN and CBR is first referenced in 1989 [43, 44], it is not until 1999 that the first true twin-systems emerge [20, 46]. From this beginning, there is a very modest development of the idea over the intervening 20 years. Indeed, citation patterns are quite inconsistent and lookbacks from more recent papers are patchy. The history divides into three distinct periods: (i) a major piece of work by a Korean Group in the late-1990s (with a parallel proposal in the USA), (ii) a significant addition by an Irish Group through the mid-2000s and, then (iii) more recent work in deep learning that revisits related approaches, often with no or poor reference to the prior literature.

### 4.1 Korean Developments (1999-2007): Feature-Weighting Tests of Twins

Around 1999, a South Korean group working at the Korean Advanced Institute of Science and Technology (KAIST), explored a range of feature-weighting techniques in comparative tests of the twin-system idea, in a framework they called "Memory Based Neural Reasoning" [46-54]. Shin and colleagues [46, 47, 51] paired MLPs with CBR operating over the same dataset, proposing that this hybrid system "can give example-based explanations together with prediction values of the neural network" [47, p.637]. Initially, they tested these twin-systems on a semiconductor-yield dataset before moving on to tests on many benchmark datasets (e.g., Iris and Wisconsin Diagnostic Breast Cancer datasets) for classification and regression tasks. In this work, they perform competitive tests of four different feature-weighting schemes for capturing the MLPs activation patterns (i.e., *sensitivity*, *activity*, *relevance* and *saliency*). For example, in *sensitivity* a feature's weight is calculated by taking the absolute difference between the normal prediction of the MLP and its prediction with that input feature set to zero; this is repeated and summed for the entire training set, and the final figure is then divided by the number of instances in the training data to normalize it for the final feature-weight value [46]. For each weighting scheme, the feature-weights were used in the $k$-NN to retrieve cases, matching the prediction for the query-case.

There are three significant contributions in this Korean work: (i) the researchers explicitly talk about a division of labor, where the ANN provides accuracy, using its feature-weights, and the CBR system provides explanations using nearest-neighbor cases, (ii) they recognize that there are many different ways to describe the ANN (i.e., different feature-weighting methods) that need to be tested[3], and (iii) they understand that there are two classes of feature-weighting methods (global and local).

Shin and colleagues [46, 47, 51] appear to be the first in the literature, to explicitly pair ANNs and CBR systems in a twinned way for purposes of explanation and to perform comparative tests of different feature-weighting schemes. Shin and colleagues [46, 47] propose that the *sensitivity* and *activity* measures seemed to perform best, (though conclusions are different for different datasets) arguing that their fidelity to the function being computed by the MLP was better. Park et al. [51] extend the earlier tests with a new feature-weighting scheme based on the important distinction between global and local weighting schemes. *Global feature-weighting* assumes the input space is isotropic, deriving a single ubiquitous feature-weight vector for the entire domain (i.e., weights do not change for different query cases), whereas *local feature-weighting* weights each specific query-case (and sometimes each training case) differently to help case retrieval. Park et al. [51] find that local-weighting schemes perform markedly better than global-weighting methods, presumably because the former captures information about a specific area of the input space for a given query in a more fine-grained way. However, their local-weighting technique is not applicable to *post-hoc* explanations in MLPs, as it is specifically trained to generate query-specific feature-weight-sets rather than giving predictions. Hence, it cannot be considered to be a twin-system; however, it does show the potential for local-weighting methods. Later work extends global weighting schemes to datasets having symbolic feature-values [53, 54]. Overall, the global methods tend to produce very similar results for different values of $k$ but the results from local methods are found to be demonstrably better.

---

[3] A fact overlooked in most papers, even very recent ones.

These nine Korean papers did not attract huge levels of attention; together they have a total of 269 GoogleScholar citations (M=30, Max=91, Min=5). Indeed, many of these citations are not specifically to the twinning idea, but reference other aspects of the work (e.g., the domains used). However, more recently, several papers have referenced their work. Weber et al. [64] claim a philosophic overlap with the Korean work in an ANN-CBR hybrid; yet the details of the feature-weighting used are not clear. Peng and Zhuang [65] propose a different feature-weighting scheme for an ANN-CBR twinning, that replaces feature-values of a case using the MLPs weights (but does not reference the Korean work). However, it is only in the last few years, that the Korean work has been seriously revisited. Biswas and colleagues [66-68] revisited the *sensitivity* measure and several limitations of earlier weighting schemes; they transform the MLP into an AND/OR graph from which weights are extracted for use in the CBR system. On applying this graph technique to several new domains, they find that it does better than other methods. Biswas et al. [68] also revisit global weighing-techniques in the context of class-imbalanced datasets, showing that a cost-sensitive learning algorithm displays improvements for such datasets. Finally, recently the importance of the global-local distinction to XAI in deep learning has been emphasized [5, 6], without referencing the Korean work.

## 4.2 A Parallel Discovery in L.A.: Caruana et al. (1999)

Around the same time as the Korean Group's work, another group working at UCLA reported an extension to an earlier system [69] that provided case-based explanations ([20] cited 33 times in GoogleScholar). Caruana et al. [20] describe a system for medical domains in which a "non-case-based learning method" (an MLP) could generate a distance metric over a training set, that could then be used to find an explanatory case that was most similar to a query-case. Caruana et al.'s MLP predicted pneumonia mortality and proposed case-based explanations based on a query-case's hidden-layer activation-vector (i.e., its latent features), by computing the Euclidean distance between the query-vector and all training-cases, thus enabling them to find the explanatory cases with the most similar latent features. The paper does not provide detailed results on the success of this method and neither does it report user trials, though it does discuss the issues surrounding how cases might be deployed to explain the ANN's predictions (for recent related work see [70]). Caruana et al.'s [20] feature-weighting method differs from those examined by the Korean group, that were based on *input space* weightings rather than *latent space* weightings (as well as being a local-weighting technique). This latent-space approach has not been pursued actively, perhaps, because it appears to be less transparent than input-space approaches (see [4]). Recently, in the deep learning literature reviews of XAI, [20] is regularly cited as *the* case-based explanation paper [7, 12, 71, 72], in the absence of references to the Korean work that, arguably, is more complete; a fact that, perhaps, indicates some discontinuities in the XAI literature.

### 4.3 An Irish Departure (mid-2000s): Local Feature-Weighting Tests of Twins

The next major step in the development of the twinning idea came in the mid-2000s, from a group of Irish researchers, largely, at University College Dublin [73-77]. This group also saw a role for CBR in explaining the opaque, but accurate outputs of MLPs, arguing that "the use of actual training data, cases from the case base, as evidence in support of a particular prediction, is a powerful and convincing form of explanation" (p. 164, [75]). The Irish researchers, who did not cite the Korean work, proposed a new and intriguing local feature-weighting method [74-76]. Nugent and Cunningham [74] were concerned with capturing the function being computed by MLPs in the local region around a given query-case for a blood-alcohol dataset. So, they systematically perturbed the features of the query-case, queried labels for these perturbed cases from the MLP, and then built a linear model from the results of these tests. The coefficients of this linear model were then used to weight the $k$-NN search in the CBR system that shared its case-base with the MLP's training set. Nugent et al. [76, 77] also considered more complex use of cases than just providing nearest neighbors, by selecting *a fortori* cases; the idea being to use a case that is closest to the decision boundary for the query-case, which may not, necessarily, be the nearest neighbor. Finally, [73] did user tests to show that the retrieved cases have some explanatory value in these domains.

There are, at least, three significant contributions from this work: it explores (i) a very different approach to the computation of local feature-weights which, in contrast to the Korean local-weighting method, can be used for *post-hoc* explanations, (ii) a more complex scheme for selecting explanatory cases, beyond the simple use of nearest neighbors, (iii) it showed that this type of twin-system had some validity for human users by using case-based explanations over feature-based ones.

These five papers have received moderate attention in the literature; between them they have a total of 236 GoogleScholar citations (M=47, max = 99, min=8). However, few of these citations are about the twin-system idea (i.e., often about user tests). Notably, though the linear-model idea has advanced significantly in the literature on interpretable classifiers [78, 79], few papers specifically cite this Irish work. For instance, the Local Interpretable Model-Agnostic Explanations (LIME) [79] technique also perturbs query-cases to build local linear models but does not cite [79]. Although, Olsson et al. [80] do, in a related approach to case similarity using logistic regression – the *principle of interchangeability* – and the notion of *local accuracy* to handle the identification of explanatory cases. More recently, there is some recognition of these papers in reviews [10, 81] but for the most part they are passed over [6, 7, 37, 72] with all CBR-solutions being attributed to Caruana et al. [20] and Kim, Rudin and Shah [21].

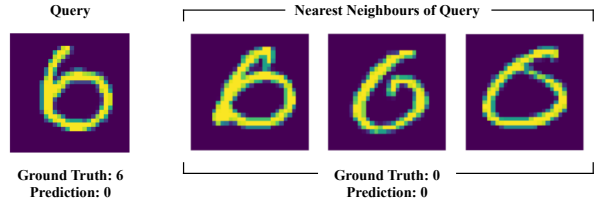

**Fig. 2.** In a CNN-CBR twin-system, the CNN classifies an image of the number "6" as "0". An explanation using nearest-neighbor cases from its CBR twin, shows the training-data used to model the CNN function in this area of the latent space was labelled as "0" but looks like "6"; so the model miss-classifies the query as it "looks like" these training cases (adapted from [4]).

### 4.4 Recent DNN-CBR Twinning

We have already established that there is a disconnect between more recent DNN research and the older twin-system literature in CBR. Yet, in the last two decades, a huge amount of work has been done on different ways to describe the functions of opaque ANN and ML systems. Recently, the focus of some of this work has shifted to the case-based explanation-by-example of DNNs. For instance, Chen et al. [82] and Li et al. [83] both build CBR into the DNN architecture itself, mainly to avoid the need for *post-hoc* explanations. Although these are not twin-systems, they do combine CBR and ANN techniques for the purpose of interpretability and explanations (though they fail to cite much previous work done). A review by Gilpin et al. [37] proposes that DNNs have often been explained by using simpler "proxy systems"[4] of which they identify four types: linear models [79], decision trees [84], automatic rule extraction [78], and saliency mapping [85-87]. Two of these approaches – linear models and saliency mapping – have resonances with the twin-systems literature reviewed here.

**Linear Models.** As in Nugent and Cunningham's work, a currently popular approach to explaining ANNs (and indeed any ML model) is to use local linear models built by perturbing an input in the neighborhood of a query. LIME [79] is *the* prime example of this approach as it finds relative feature-weightings for a given query-case. Recently, LIME has been used in comparative tests of several twin-systems ([4] influenced by Nugent and Cunningham) and found that it not appreciably better than popular global-weighting techniques (including the *sensitivity* method used by the Korean group).

**Saliency Mapping.** Another popular technique looks at the contribution of inputs to an ANN's output, deriving saliency maps by backpropagating contribution scores from a given activation in the network (usually in the output layer), to a previous layer (usually the input one). Amongst others, these methods include *Layer-wise Relevance Propagation* [85], *Integrated Gradients* [86], and *DeepLIFT* [87]. This *saliency mapping* is typically used to highlight important pixels in a CNN's classification of an image, however it has other uses and has recently been applied to MLP-CBR and CNN-CBR twin-systems (notably, with multiple fully connected dense layers using image transfer learning [88]) in comparative tests (see [4] and also Fig 2).

---

[4]   The proxy system is meant to behave similarly to the black-box system but is simpler for explanatory purposes (so, the CBR in ANN-CBR twin is one type of proxy model).

**Other DNN Options.** There are also a handful of other DNN options that are arguably twin-systems though of quite a different ilk. Work on the extraction of prototypes from the analysis of DNNs have been cast as case-based approaches, though with a Bayesian aspect [21, 83]. These proposals look like a different type of twinning – Bayesian-CBR twins – that perhaps have other precursors in the CBR literature [89, 90]. Another approach tries to map the layers of a DNN onto particular exemplar cases using Deep *k*-Nearest neighbors (DkNN) [71, 91, 92]. However, it still remains to be seen whether these are to be accommodated as twin-systems.

## 5 Future Directions: Road Mapping

In the present paper, we have reviewed the history of how CBR has been used in a twinning fashion to explain the outputs of ANNs. The significance and importance of this survey is that it shows there are generalizations about XAI to be gleaned from the twin-system approach. Such generalizations may help us avoid the current scattered fragmentary development of XAI solutions [5]. This review also suggests a research road map for future work in this area, along at least three paths:

- *Feature-Weighting Schemes*. It is clear that there is a large space of feature-weighting schemes that could be explored (especially, more recent ones in DNNs); this exploration needs to be done in a controlled and comparative fashion to determine which ones are best for which domains and tasks.
- *The Deployment of Cases*. CBR work has shown in twin-systems there are many different ways cases can be used for explanation (e.g., *a fortiori* usage, counterfactual cases, near misses, nearest unlike neighbors and so on; more needs to be done on these ideas in the context of ANNs, and especially DNNs.
- *The Embarrassment of User Testing.* In all the papers we examined we found less than a handful (i.e., < 5) that performed any adequate user testing of the proposal that cases improved the interpretability of models; this gap needs to be rectified.

In conclusion, notwithstanding the citation gaps in the literature, it is clear that there are many fruitful directions in which the CBR-twin idea can be taken to answer the interpretability problems we currently face in XAI.

### References

1. Adadi, A. and Berrada, M.: Peeking inside the black-box: A survey on Explainable Artificial Intelligence (XAI). IEEE Access, 6, 52138-52160 (2018)
2. Goodman, B. and Flaxman, S.: European Union regulations on algorithmic decision-making and a "right to explanation". AI Magazine, 38(3), 50-57 (2017)
3. Wachter, S., Mittelstadt, B. and Floridi, L.: Why a right to explanation of automated decision-making does not exist in the general data protection regulation. International Data Privacy Law, 7(2), 76-99 (2017)
4. Kenny, E.M. and Keane, Mark T.: Twin-systems for explaining ANNs using CBR: Experiments using feature-weighting schemes to retrieve explanatory cases. Under review (2019)

5.  Guidotti, R., Monreale, A., Ruggieri, S., Turini, F., Giannotti, F. and Pedreschi, D.: A survey of methods for explaining black box models. ACM Computing Surveys, 51(5), p.93 (2018)
6.  Doshi-Velez, F. and Kim, B.: Towards a rigorous science of interpretable machine learning. arXiv preprint arXiv:1702.08608 (2017)
7.  Lipton, Z. C.: The mythos of model interpretability. Queue, 16(3), 30 (2018)
8.  Miller, T.: Explanation in artificial intelligence: Insights from the social sciences. Artificial Intelligence, 267, 1-38 (2019)
9.  Abdul, A., Vermeulen, J., Wang, D., Lim, B.Y. and Kankanhalli, M.: Trends and trajectories for explainable, accountable and intelligible systems: An HCI research agenda. In: Proceedings 2018 CHI Conference on Human Factors in Computing Systems (p. 582). ACM (2018)
10. Biran, O. and Cotton, C.: Explanation and justification in machine learning: A survey. In: IJCAI-17 workshop on explainable AI (XAI) (Vol. 8) p.1 (2017)
11. Sørmo, F., Cassens, J. and Aamodt, A.: Explanation in case-based reasoning–perspectives and goals. Artificial Intelligence Review, 24(2), 109-143 (2005)
12. Johs, A.J., Lutts, M. and Weber, R.O.: Measuring Explanation Quality in XCBR. ICCBR18, p.75 (2018)
13. Tintarev, N. and Masthoff, J.: A survey of explanations in recommender systems. In: 2007 IEEE 23rd international conference on data engineering workshop, 801-810. IEEE (2007)
14. Harman, G.H.: The inference to the best explanation. The philosophical review, 74(1), 88-95 (1965)
15. Salmon, W.C.: Scientific explanation and the causal structure of the world. Princeton University Press (1984)
16. Van Fraassen, B.C.: The scientific image. Oxford University Press (1980)
17. Keil, F.C.: Explanation and understanding. Annual Rev. of Psychology, 57, 227-254 (2006)
18. Leake, D.B.: CBR in context: The present and future. Case-based reasoning: Experiences, lessons, and future directions, pp.3-30 (1996)
19. Leake, D. and McSherry, D.: Introduction to the special issue on explanation in case-based reasoning. Artificial Intelligence Review, 24(2), 103-108 (2005)
20. Caruana, R., Kangarloo, H., Dionisio, J.D., Sinha, U. and Johnson, D.: Case-based explanation of non-case-based learning methods. In: Proceedings of the AMIA Symposium, p. 212. American Medical Informatics Association (1999)
21. Kim, B., Rudin, C. and Shah, J.A.: The Bayesian case model: A generative approach for case-based reasoning and prototype classification. In: Adv. in NIPs, pp. 1952-1960 (2014)
22. Pedreschi, D., Giannotti, F., Guidotti, R., Monreale, A., Ruggieri, S. and Turini, F.: Meaningful explanations of Black Box AI decision systems. In: Proceedings of AAAI-19 (2019)
23. Haykin, S.: Neural networks (Vol. 2). New York: Prentice Hall (1994)
24. Goodfellow, I., Bengio, Y. and Courville, A.: Deep learning. MIT Press (2016)
25. Kolodner, J.: Case based reasoning. Morgan Kaufmann (2014)
26. Aamodt, A. and Plaza, E.: Case-based reasoning: Foundational issues, methodological variations, and system approaches. AI communications, 7(1), 39-59 (1994)
27. De Mantaras, R.L., McSherry, D., Bridge, D., Leake, D., Smyth, B., Craw, S., Faltings, B., Maher, M.L., Cox, M.T., Forbus, K. and Keane, M.T.: Retrieval, reuse, revise and retention in CBR. Knowledge Engineering Review, 20(3), 215-240 (2006)
28. Sahin, S., Tolun, M.R. and Hassanpour, R.: Hybrid expert systems: A survey of current approaches and applications. Expert systems with applications, 39(4), 4609-4617 (2012)
29. Negnevitsky, M.: Artificial intelligence. Pearson education (2005)
30. Medsker, L.R.: Hybrid neural network and expert systems. Springer (2012)
31. Szegedy, C., Zaremba, W., Sutskever, I., Bruna, J., Erhan, D., Goodfellow, I. and Fergus, R.: Intriguing properties of neural networks. arXiv preprint arXiv:1312.6199 (2013)

32. Krizhevsky, A., Sutskever, I. and Hinton, G.E.: Imagenet classification with deep convolutional neural networks. In: Advances in NIPs, pp. 1097-1105 (2012)
33. Witten, Ian H., Eibe Frank, Mark A. Hall, and Christopher J. Pal.: Data Mining: Practical machine learning tools and techniques. Morgan Kaufmann (2016)
34. LeCun, Y., Bengio, Y. and Hinton, G.: Deep learning. Nature, 521(7553), p.436 (2015)
35. Olden, J.D. and Jackson, D.A.: Illuminating the "black box". Ecological modelling, 154(1-2), pp.135-150 (2002)
36. Selvaraju, R.R., Cogswell, M., Das, A., Vedantam, R., Parikh, D. and Batra, D.: Grad-cam: Visual explanations from deep networks via gradient-based localization. In: Proceedings of the IEEE International Conference on Computer Vision. pp. 618-626 (2017)
37. Gilpin, L.H., Bau, D., Yuan, B.Z., Bajwa, A., Specter, M. and Kagal, L.: Explaining explanations: an approach to evaluating interpretability of machine learning. arXiv preprint arXiv:1806.00069 (2018)
38. Zeiler, M.D. and Fergus, R.: Visualizing and understanding convolutional networks. In: European conference on computer vision, pp. 818-833. Springer (2014)
39. He, Kaiming, Xiangyu Zhang, Shaoqing Ren, and Jian Sun: Deep residual learning for image recognition. In: Proceedings of the IEEE conference on computer vision and pattern recognition, pp. 770-778 (2016)
40. Erhan, D., Bengio, Y., Courville, A. and Vincent, P.: Visualizing higher-layer features of a deep network. University of Montreal, 1341(3), p.1 (2009)
41. Keane, M.T.: Analogical asides on case-based reasoning. In: European Workshop on Case-Based Reasoning, pp. 21-32, Springer, Berlin, Heidelberg (1993)
42. Nunes, I. and Jannach, D.: A systematic review and taxonomy of explanations in decision support and recommender systems. UMUAI, 27(3-5), 393-444 (2017)
43. Becker, L. and Jazayeri, K.: A connectionist approach to case-based reasoning. In: Proceedings of the Case-Based Reasoning Workshop, pp. 213-217. Morgan Kaufmann (1989)
44. Thrift, P.: A neural network model for case-based reasoning. In: Proceedings of the Case-Based Reasoning Workshop, pp. 334-337. Morgan Kaufmann (1989)
45. Hilario, M., Pellegrini, C., & Alexandre, F. Modular integration of connectionist and symbolic processing in knowledge-based systems. C.de R. en Informatique de Nancy (1994)
46. Shin, C. K., & Park, S. C.: Memory and neural network based expert system. Expert Systems with Applications, 16(2), 145-155 (1999)
47. Shin, C. K., Yun, U. T., Kim, H. K., & Park, S. C.: A hybrid approach of neural network and memory-based learning to data mining. IEEE Transactions on Neural Networks, 11(3), 637-646 (2000)
48. Shin, C.K. and Park, S.C., A machine learning approach to yield management in semiconductor manufacturing. International Journal of Production Research, 38, 4261-4271 (2000)
49. Park, J.H., Shin, C.K., Im, K.H. and Park, S.C.: A local weighting method to the integration of neural network and case based reasoning. In: Neural Networks for Signal Processing XI: Proceedings of the 2001 IEEE SPS Workshop, pp. 33-42. IEEE (2001)
50. Shin, C.K. and Park, S.C.: Towards integration of memory based learning and neural networks. In Soft computing in case based reasoning (pp. 95-114). Springer, London (2001)
51. Park, J.H., Im, K.H., Shin, C.K. and Park, S.C.: MBNR: case-based reasoning with local feature weighting by neural network. Applied Intelligence, 21(3), pp.265-276 (2004)
52. Park, S.C., Kim, J.W. and Im, K.H.: Feature-weighted CBR with NN for symbolic features. In: International Conference on Intelligent Computing, 1012-1020. Springer (2006)
53. Im, H. and Park, S.C.: Case-based reasoning and neural network based expert system for personalization. Expert Systems with Applications, 32(1), pp.77-85 (2007)

54. Ha, S.: A personalized counseling system using case-based reasoning with neural symbolic feature weighting (CANSY). *Applied intelligence*, *29*(3), pp. 279-288 (2008)
55. Reategui, E.B., Campbell, J.A. and Leao, B.F.: Combining a neural network with case-based reasoning in a diagnostic system. Artificial Intelligence in Medicine, 9(1), 5-27 (1997)
56. Yang, B.S., Han, T. and Kim, Y.S.: Integration of ART-Kohonen NN and CBR for intelligent fault diagnosis. Expert Systems with Applications, 26(3), 387-395 (2004)
57. Rodriguez, Y., Garcia, M.M., De Baets, B., Morell, C. and Bello, R.: A connectionist fuzzy case-based reasoning model. In: Mexican International Conference on Artificial Intelligence, pp. 176-185. Springer, Berlin, Heidelberg (2006)
58. Amin, K., Kapetanakis, S., Althoff, K.D., Dengel, A. and Petridis, M.: Answering with cases: A CBR Approach to Deep Learning. In: International Conference on Case-Based Reasoning, pp. 15-27. Springer, Cham. (2018)
59. Corchado, J.M., Rees, N., Lees, B. and Aiken, J.: Data mining using example-based methods in oceanographic forecast models. In: IEE Colloquium on Knowledge Discovery and Data Mining (Digest No. 1998/310) (pp. 7-1). IET (1998)
60. Corchado, J.M. and Lees, B.: A hybrid case-based model for forecasting. Applied Artificial Intelligence, 15(2), 105-127 (2001)
61. Fdez-Riverola, F., Corchado, J.M. and Torres, J.M.: An automated hybrid CBR system for forecasting. In: European Conf. on Case-Based Reasoning, pp. 519-533. Springer (2002)
62. Jothikimar, R., Shivakumar, N., Ramesh, P.S., Suganthan, and Suresh, A.: Heart Disease Prediction System Using ANN, RBF and CBR. International Journal of Pure and Applied Mathematics, 117(21), 199-217 (2017)
63. Kouser, R.R., Manikandan, T. and Kumar, V.V.: Heart Disease Prediction System Using Artificial Neural Network, Radial Basis Function and Case Based Reasoning. Journal of Computational and Theoretical Nanoscience, 15(9-10), 2810-2817 (2018)
64. Weber, R., Proctor, J.M., Waldstein, I. and Kriete, A.: CBR for modeling complex systems. In: International Conference on Case-Based Reasoning, pp. 625-639. Springer (2005)
65. Peng, Y. and Zhuang, L.: A case-based reasoning with feature weights derived by BP network. In: Intelligent Information Technology Application, pp. 26-29. IEEE (2007)
66. Biswas, S. K., Sinha, N., Purakayastha, & Marbaniang, L.: Hybrid expert system using case based reasoning and neural network for classification. Biologically Inspired Cognitive Architectures, 9, 57-70 (2014)
67. Biswas, S. K., Baruah, B., Sinha, N., & Purkayastha, B.: A hybrid CBR classification model by integrating ANN into CBR. International Journal of Services Technology and Management, 21(4-6), 272-293 (2015)
68. Biswas, S. K., Chakraborty, M., Singh, H. R., Devi, D., Purkayastha, B., & Das, A. K.: Hybrid case-based reasoning system by cost-sensitive neural network for classification. Soft Computing, 21(24), 7579-7596 (2017)
69. Cooper, G.F., Aliferis, C.F., Ambrosino, R., Aronis, J., Buchanan, B.G., Caruana, R., Fine, M.J., Glymour, C., Gordon, G., Hanusa, B.H. and Janosky, J.E.: An evaluation of machine-learning methods for predicting pneumonia mortality. Artificial intelligence in medicine, 9(2), 107-138 (1997)
70. Caruana, R., Lou, Y., Gehrke, J., Koch, P., Sturm, M. and Elhadad, N.: Intelligible models for healthcare: Predicting pneumonia risk and hospital 30-day readmission. In: Proceedings of the 21th ACM SIGKDD International Conference on Knowledge Discovery and Data Mining, pp. 1721-1730. ACM (2015)
71. Papernot, N., & McDaniel, P. Deep k-nearest neighbours: Towards confident, interpretable and robust deep learning. arXiv preprint arXiv:1803.04765 (2018)

72. Mittelstadt, B., Russell, C. and Wachter, S.: Explaining Explanations in AI. In: Proceedings of Conference on Fairness, Accountability, and Transparency (FAT*-19) (2019)

73. Cunningham, P., Doyle, D., and Loughrey, J.: An evaluation of the usefulness of case-based explanation. In: International Conf. on Case-Based Reasoning, 122-130. Springer (2003)

74. Nugent, C., and Cunningham, P.: A case-based explanation system for black-box systems. Artificial Intelligence Review, 24(2), 163-178 (2005)

75. Nugent, C., Cunningham, P., and Doyle, D.: The best way to instill confidence is by being right. In: International Conference on Case-Based Reasoning, pp. 368-381. Springer (2005)

76. Nugent, C., Doyle, D., and Cunningham, P.: Gaining insight through case-based explanation. Journal of Intelligent Information Systems, 32(3), 267-295 (2009)

77. Doyle, D., Cunningham, P., Bridge, D. and Rahman, Y.: Explanation oriented retrieval. In European Conference on Case-Based Reasoning, pp. 157-168. Springer (2004)

78. Andrews, R., Diederich, J. and Tickle, A.B.: Survey and critique of techniques for extracting rules from trained artificial neural networks. Knowledge-based systems, 8, 373-389 (1995)

79. Ribeiro, M.T., Singh, S. and Guestrin, C.: Why should I trust you?: Explaining the predictions of any classifier. In: Proceedings of the 22nd ACM SIGKDD international conference on knowledge discovery and data mining, pp. 1135-1144. ACM (2016)

80. Olsson, T., Gillblad, D., Funk, P. and Xiong, N.: Case-based reasoning for explaining probabilistic machine learning. International Journal of Computer Science and Information Technology, 6(2), 87-101 (2014)

81. Zharov, Y., Korzhenkov, D., Shvechikov, P. and Tuzhilin, A.: YASENN: Explaining Neural Networks via Partitioning Activation Sequences. arXiv preprint arXiv:1811.02783 (2018)

82. Chen, C., Li, O., Barnett, A., Su, J. and Rudin, C.: This looks like that: deep learning for interpretable image recognition. *arXiv preprint arXiv:1806.10574* (2018)

83. Li, O., Liu, H., Chen, C., and Rudin, C.: Deep learning for case-based reasoning through prototypes: A neural network that explains its predictions. In: Thirty-Second AAAI Conference on Artificial Intelligence. AAAI (2018)

84. Zilke, J.R., Mencía, E.L. and Janssen, F.: DeepRED–Rule extraction from deep neural networks. In: Internat. Conf. on Discovery Science, pp. 457-473. Springer (2016)

85. Bach, S., Binder, A., Montavon, G., Klauschen, F., Müller, K.R. and Samek, W.: On pixel-wise explanations for non-linear classifier decisions by layer-wise relevance propagation. PloS one, 10(7), p.e0130140. (2015)

86. Sundararajan, M., Taly, A. and Yan, Q.: Axiomatic attribution for deep networks. In: Proceedings of the 34th International Conference on Machine Learning-Volume, 70, 3319-3328. JMLR. Org (2017)

87. Shrikumar, A., Greenside, P. and Kundaje, A.: Learning important features through propagating activation differences. In: Proceedings of the 34th International Conference on Machine Learning-Volume 70, 3145-3153. JMLR. Org (2017)

88. Zhang, C.L., Luo, J.H., Wei, X.S. and Wu, J.: In defense of fully connected layers in visual representation transfer. In: *Pacific Rim Conf. on Multimedia* (pp. 807-817). Springer (2017)

89. Myllymäki, P. and Tirri, H.: November. Massively parallel case-based reasoning with probabilistic similarity metrics. In: European Workshop on Case-Based Reasoning (pp. 144-154). Springer, Berlin, Heidelberg (1993)

90. Kofod-Petersen, A., Langseth, H., and Aamodt, A.: Explanations in Bayesian networks using provenance through case-based reasoning. In: Workshop Proceedings, p. 79 (2010)

91. Wallace, E., Feng, S. and Boyd-Graber, J.: Interpreting Neural Networks with Nearest Neighbours. arXiv preprint arXiv:1809.02847 (2018)

92. Card, D., Zhang, M. and Smith, N.A. January. Deep Weighted Averaging Classifiers. In: Proceedings of Conf. on Fairness, Accountability & Trust (pp. 369-378). ACM (2019)